\pdfoutput=1

\documentclass[11pt]{article}

\usepackage[final]{acl}
\usepackage{booktabs}
\usepackage{tabularx}
\usepackage{listings}
\usepackage{times}
\usepackage{latexsym}

\usepackage[T1]{fontenc}

\usepackage[utf8]{inputenc}

\usepackage{microtype}

\usepackage{inconsolata}

\usepackage{graphicx}
\usepackage{subfigure}
\usepackage{booktabs}

\usepackage{amsmath}
\usepackage{amsthm}
\usepackage{enumitem}
\usepackage{color,amsfonts}
\usepackage{makecell}
\usepackage{multirow,array}
\usepackage{tikz}
\usepackage{pgfplots}
\usepackage{amssymb,mathtools}
\usepackage[linesnumbered,ruled,vlined]{algorithm2e}
\usepackage{caption}
\usepackage{cleveref}
\usepackage[table]{xcolor}
\usepackage{arydshln}
\usepackage{amssymb}

\newcommand{\model}[1]{\textsc{#1}\xspace}
\newcommand{\ours}{\model{TransGraph}}

\title{Discourse Graph Guided Document Translation with Large Language Models}

\author{Viet-Thanh Pham, Minghan Wang, Hao-Han Liao, Thuy-Trang Vu \\
  Department of Data Science \& AI, Monash University \\ 
  \texttt{\{thanh.pham1,minghan.wang,hlia0034@student,trang.vu1\}@monash.edu}
}

\begin{document}
\maketitle
\begin{abstract}
Adapting large language models to full document translation remains challenging due to the difficulty of capturing long-range dependencies and preserving discourse coherence throughout extended texts. 
While recent agentic machine translation systems mitigate context window constraints through multi-agent orchestration and persistent memory, they require substantial computational resources and are sensitive to memory retrieval strategies. 
We introduce \ours, a discourse-guided framework that explicitly models inter-chunk relationships through structured discourse graphs and selectively conditions each translation segment on relevant graph neighbourhoods rather than relying on sequential or exhaustive context.
Across three document-level MT benchmarks spanning six languages and diverse domains, \ours consistently surpasses strong baselines in translation quality and terminology consistency while incurring significantly lower token overhead.
\end{abstract}
\section{Introduction}






Recent advances in large language models (LLMs) have revolutionised machine translation (MT). LLM-based MT models have achieved remarkable performance at the sentence level and demonstrated promising capabilities for document-level translation (DocMT), often surpassing specialised encoder-decoder MT systems~\citep{xu2024a,lyu-etal-2024-paradigm,wu2024adapting,pang2025salute}. Despite these successes, translating entire long documents remains a challenge, particularly in the modelling of long-range dependencies and discourse phenomena across sentences and paragraphs.

Fine-tuning LLMs on high-quality document-level data has improved multilingual contextualisation, yet the limited context window poses a persistent bottleneck~\citep{li-etal-2025-enhancing-large,ramos2025multilingual}.  Recent agentic MT approaches address this by coordinating multiple specialised LLM agents with persistent memory to store and retrieve discourse information, such as proper nouns, bilingual summaries and short/long-term cache, throughout translation~\citep{transagents_demo,transagent,delta_arxiv,graft2025}.
These systems achieve impressive quality and style control, including for literary text, but often at high orchestration and token costs, with quality sensitive to memory hygiene and retrieval placement. 
Meanwhile, other research suggests that structured and selective context---rather than \emph{more} context---can be key: well-chosen cross-sentential signals and explicit discourse cues (coreference, cohesion devices) can produce document-level gains without complex pipelines \citep{zhang2022rethinking,codonmt2022}.

In this paper, we take a discourse-guided, \emph{selective-context} approach and propose \emph{\ours}, a two-stage, graph-conditioned procedure for document-level MT. Rather than relying on exhaustive sequential context or memory-intensive retrieval systems, we first partition a document into coherent chunks and construct a labelled inter-chunk graph that explicitly encodes discourse relations frequently driving translation choices, such as \emph{Entity-Coreference}, \emph{Core\(\rightarrow\)Detail}, \emph{Motivation\(\rightarrow\)Method}, drawing on Rhetorical Structure Theory and modern discourse parsing insights \citep{mann1988rst,taboada2006rst,jiang2023discourse}.
During translation, each chunk is conditioned not on all preceding text, but on a small, discourse-guided neighbourhood, enriched with discourse relation labels.
\ours offers three key advantages: (i) selective conditioning only on the contextual information that matters for the translation; (ii) improving terminology and cohesion with minimal token overhead; and (iii) remaining fully agnostic to the underlying model or backbone architecture.

\begin{figure*}[t]    
\includegraphics[width=0.99\linewidth]{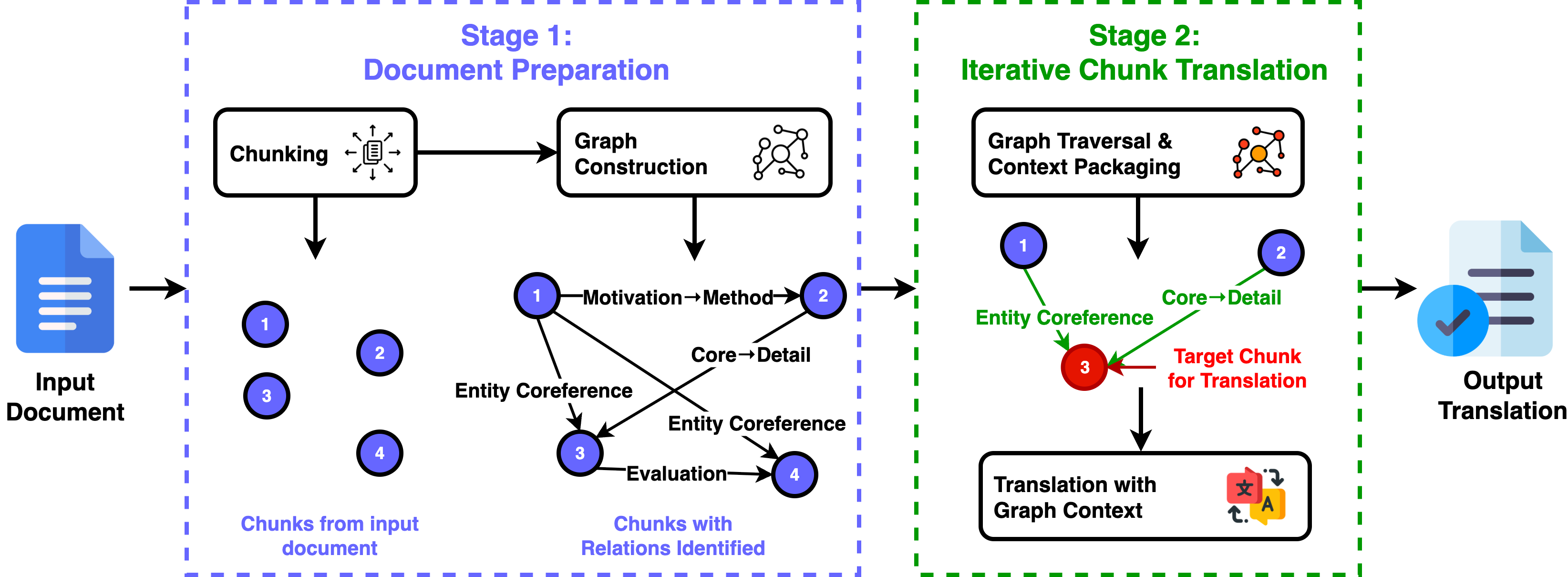}
\caption{Illustration of our proposed framework, \emph{TransGraph}. In Stage 1, \emph{TransGraph} takes a document as input, split it into multiple small chunks and identify the discourse relations between every pair of chunks and represent them as a knowledge graph. In Stage 2, the translation of one chunk proceeds by retrieving adjacent chunks in the graph with their corresponding relations.}
\label{fig:pipeline}
\end{figure*}

We evaluate \ours on three DocMT benchmarks featuring tagged terminologies across diverse domains, including technical presentations and literary texts, spanning six languages. 
Our method consistently outperforms sentence-level baselines, sequential-context DocMT, and agentic MT systems in both document-level BLEU and COMET scores, as well as terminology accuracy while incurring substantially lower token costs.
In summary, our contributions are:

\begin{itemize}[noitemsep,topsep=2pt,parsep=2pt,partopsep=2pt]
  \item A discourse graph guided DocMT framework that injects \emph{relation-labelled} document context, aligning with discourse theory while remaining lightweight and backbone-agnostic.
  \item We demonstrate that structured discourse relations outperform generic sequential context expansion, achieving consistent improvements in translation quality.
  \item Extensive comparisons to agentic MT systems, highlighting a strong quality--cost trade-off and robustness under long documents.
\end{itemize}

\section{Document Translation with Discourse Graph}









\label{sec:method}

We present \emph{\ours}, a two-stage, graph-conditioned procedure for document-level machine translation, illustrated in \Cref{fig:pipeline}. In Stage 1, \emph{\ours} first splits the source document into coherent textual units (chunks) and construct a directed graph of discourse-level relations between chunks. In Stage 2, the framework then performs translation while conditioning each chunk on graph-retrieved context. We denote \(V=\{c_1,\ldots,c_N\}\) the set of chunks and \(G=(V,E,\ell)\) a directed, labeled graph representing the discourse relations among pairs of chunks, where \(E\subseteq V\times V\) and \(\ell:E\to\mathcal{R}\) assigns a label for an edge from a predefined schema, capturing discourse-level relations such as \emph{Entity-Coreference}, \emph{Core\(\rightarrow\)Detail}, \emph{Motivation\(\rightarrow\)Method}, and \emph{Evaluation}. Below sections will discuss each step in the procedure.

\subsection{Stage 1: Document Preparation}
\label{subsec:stage1}

The objective of Stage~1 is to produce coherent chunks and especially an interpretable discourse graph that captures salient inter-chunk relations within a bounded window, thereby providing structured context for downstream translation.
\vspace{-1.5mm}
\paragraph{Chunking}
The document is segmented into contiguous chunks that preserve sentence boundaries. We iteratively loop through \(T\) tokens of the source document and prompt the LLM to split the text into chunks (Prompt template shown in \Cref{app:prompt_templates}, \Cref{listing-chunk-prompt}). These chunks are then appended to the chunk set \(V\) and the LLM continues to process the next \(T\) tokens. In case the last \(n\) tokens of \(T\) do not end with a punctuation, we merge them with the next \(T\) tokens and ask LLM to perform chunking again. This process ensures that (i) we avoid the long-context problem of LLM by providing it with limited \(T\) tokens and (ii) we preserve strict sentence boundaries.
\vspace{-1.5mm}
\paragraph{Graph Construction and Relation Labeling}
Inter-chunk structure is represented as a directed, labeled graph \(G=(V,E,\ell)\). To control computational cost while preserving locality, we only identify relations of forward pairs within a small window  \(\mathcal{P}\):
$
\mathcal{P}=\{(i,j): 1\le i<j\le N,\; j-i\le w\}.
$
where \(w\) is the window size, \((i,j)\) are the indexes of chunk \(i\) and \(j\). For each \((i,j)\in\mathcal{P}\), the LLM is prompted with the chunk pair \((c_i,c_j)\) to decide whether a meaningful relation holds, its direction, and a label \(\ell(c_i,c_j)\in\mathcal{R}\) for each edge (Prompt template in \Cref{app:prompt_templates}, \Cref{listing-relation-identification}). 

Based on prior works and theories on rhetorical structure \citep{mann1988rst,taboada2006rst,jiang2023discourse}, we define 10 types of relation for the LLM to choose from. Besides several existing types of relation (e.g. \emph{Motivation\(\rightarrow\)Method}, \emph{Cause\(\rightarrow\)Effect}, and \emph{Contrast}), we also added several relation types focusing on ensuring coherent translation and terminology consistency, such as \emph{Entity-Coreference} and \emph{Terminology Definition}. All relation types in our setup are described in \Cref{tab:chunk_relations}. Pairs judged as \emph{No-Relation} are discarded. 

\begin{table*}[htbp]
\footnotesize
\centering
\label{tab:chunk_relations}
\resizebox{0.99\textwidth}{!}{%
\begin{tabular}{llp{12cm}}
\toprule
\textbf{Relation (value)} & \textbf{Typical direction} & \textbf{Definition} \\
\midrule
Background$\rightarrow$Core & forward & $i$ supplies necessary context that enables understanding claims or results in $j$.\\
\midrule
Core$\rightarrow$Detail & forward & $j$ elaborates on or narrows a point introduced in $i$ (part–whole, attribute, instance). \\
\midrule
Problem$\rightarrow$Solution & forward & $i$ states a need, gap, or problem; $j$ presents a method or solution addressing it.\\
\midrule
Cause$\rightarrow$Effect & forward & $i$ presents a cause or reason; $j$ describes the resulting outcome.\\
\midrule
Contrast & either (often symmetric) & $i$ and $j$ assert opposing or divergent claims, properties, or outcomes. \\
\midrule
Comparison & either (often symmetric) & $i$ and $j$ compare entities along shared dimensions without opposition.\\
\midrule
Condition & forward (common) & $i$ states a condition or hypothesis; $j$ describes what holds under that condition. \\
\midrule
Evaluation & forward & $j$ evaluates, critiques, or validates the approach introduced in $i$.\\
\midrule
Entity Coreference & either (symmetric) & $i$ and $j$ mention the same entity (dataset, model, variable) establishing continuity.\\
\midrule
Terminology Definition & forward (common) & $i$ defines a term that appears or is used in $j$.\\
\midrule
\textit{No relation} & N/A & No meaningful topical or semantic connection between $i$ and $j$.\\
\bottomrule
\end{tabular}
}
\caption{Discourse-level relations used for knowledge graph construction. Direction: \emph{forward} means earlier Chunk $i \rightarrow$ later Chunk $j$; \emph{either} is either Chunk $i \rightarrow$  Chunk $j$ or Chunk $j \rightarrow$  Chunk $i$ are acceptable.}
\vspace{-1em}
\end{table*}

\subsection{Stage 2: Iterative Chunk Translation}
\label{subsec:stage2}

Given \((V,G)\), translation proceeds by traversing the graph, assembling compact context for each target chunk from its in-neighborhood, and prompting the LLM to produce the target-language translation with the assembled context. The stage is iterative in the sense that the procedure is repeated for each chunk. By conditioning each translation step on the appropriate discourse relations to neighboring contents, the system improves referent resolution, terminology consistency, and rhetorical coherence.

\vspace{-1.5mm}
\paragraph{Graph Traversal}

Chunks are processed in a stable index order, consistent with the document flow. For a target chunk \(c_j\) for translation, we gather its immediate in-neighbors in the graph
$
\mathcal{N}^{-}(j)=\{\,i\;|\;(c_i,c_j)\in E\,\}.
$
This choice aligns with the causal direction of discourse found in natural documents: antecedents typically precede anaphors, motivations precede methods, and core statements precede elaborations \cite{mann1988rst}, therefore leading to better consistency in translation.

\vspace{-1.5mm}
\paragraph{Context Packaging}

For each chunk index \(j\), we assemble a structured context package \(\mathcal{C}_j\) comprised of a small set of formatted records from \(\mathcal{N}^{-}(j)\):
$
\mathcal{C}_j=\{(i,\,c_i,\,\ell(c_i,c_j)):\; i\in\mathcal{N}^{-}(j)\}.
$
Each record includes the neighbor chunk content and its relation label, enabling the translator to treat evidence according to its discourse role rather than as undifferentiated context (Example provided in Appendix). To avoid the long-context problem of LLMs, \(|\mathcal{N}^{-}(j)|\) is capped; ties are resolved by preferring earlier neighbors (e.g. pair of chunks that are adjacent to each other), which empirically benefits term consistency and document coherence.

\vspace{-1.5mm}
\paragraph{Translation with Graph Context}

The translation of \(c_j\) is produced by prompting an LLM with the pair \((c_j,\mathcal{C}_j)\). The prompt (given in \Cref{app:prompt_templates}, \Cref{listing-chunk-translation}) instructs the model to preserve terminology and discourse relation specified by each relation label \(\ell(c_i,c_j)\). Denoting the translation operator realized by the LLM as \(T(\cdot)\), the output is represented as
$\hat{y}_j = T\!\left(c_j \,\middle|\, \mathcal{C}_j\right).
$
After all chunks are translated, the target document is reconstructed by concatenation in source order:
$\hat{Y}=\hat{y}_1 \,\Vert\, \cdots \,\Vert\, \hat{y}_N.$
Because each translation is conditioned on its in-neighborhood, boundary artifacts are mitigated without imposing expensive global decoding constraints.

\section{Experiments}

\subsection{Experimental Setups}

We conduct experiments with different open-source families of LLMs, including Qwen3~\citep{qwen3}, Llama3~\citep{grattafiori2024llama} and Ministral\footnote{Model signatures: \texttt{Qwen/Qwen3-8B}, \texttt{Qwen/Qwen3-14B}, \texttt{Qwen/Qwen3-32B}, \texttt{meta-llama/Llama-3.1-8B-Instruct} and \texttt{mistralai/Ministral-8B-Instruct-2410}.}. 
Decoding parameters, such as temperature, top-k, for each model are set to the recommended settings from the authors of these LLMs. Regarding the hyperparameter choices,
we empirically set $T$ in Stage 1 (Chunking) to be 100 tokens, while the number of in-neighbour chunks $\mathcal{N}^{-}(j)$ to retrieve in Stage 2 is set to 5 chunks.


To demonstrate the performance and efficiency of \textsc{TransGraph}, we compare our proposed framework with the following baselines:
\begin{itemize}[noitemsep,topsep=2pt,parsep=2pt,partopsep=2pt]
\item\textbf{Sentence-Level Translation (\textsc{Sent. MT})} We prompt LLM to iteratively translate each sentence in the source documents.
\item\textbf{Single-Pass Document Translation (\textsc{1-Pass DocMT)}} We prompt LLM to translate the whole source document in a single run.
\item\textbf{TransAgent \citep{transagent}} - a multi-agent framework designed for document translation. It incorporates several steps to help LLMs maintain consistency while translating long-form text, such as summarization, glossary construction, and tone and style detection.
\item\textbf{DeLTA \citep{delta_arxiv}} - an agentic framework that focus on maintaining translation consistency. It features a multi-level memory structure that stores information across various granularities and spans, including proper noun records, bilingual summary,
long-term memory, and short-term memory, which are continuously retrieved
and updated by auxiliary LLM-based components.


\item\textbf{\ours - Fixed Size Chunking (\textsc{Fixed})} We remove the chunking component of TransGraph and use fixed-size chunking. Specifically, for a given input document, we split the document into 10 length-equivalent chunks, based on the number of tokens.
\item\textbf{\ours - Removing Discourse Relations (\textsc{$-$Rel})} Instead of iteratively translate chunks conditioning on the neighboring chunks and their discourse relations, we take 5 previous chunk contents as context for the translation of the current chunk.
\item\textbf{\ours - Conditioning on Sequential Context (\textsc{Seq})} We modify the Context Packaging step of \textsc{TransGraph}, in which we only select chunk $i$ for translating the current chunk $j$ only if $i$ is in the range $[j-1, j-5]$.
\end{itemize}


\vspace{-1.5mm}
\paragraph{Benchmarks}

To assess the performance of translation methods, we select three document translation benchmarks. Among these three, two benchmarks have tagged terms in both source and target documents, from which we can calculate the accuracy of translation methods in maintaining terminology accuracy. \Cref{tab:dataset-stats} shows the statistics of each dataset.

\begin{table}[t]
\centering
\footnotesize
\resizebox{0.99\columnwidth}{!}{
\begin{tabular}{lrrr}
\toprule
\textbf{Dataset} & \textbf{ACL 60/60} & \textbf{BWB} & \textbf{GuoFeng} \\
\midrule
Total \# of Docs        & 5      & 80    & 25    \\
Avg \# of Sentences     & 84.20  & 39.91 & 57.86 \\
Max \# of Sentences     & 101    & 47    & 143   \\
Min \# of Sentences     & 57     & 19    & 28    \\
Avg \# of Tokens        & 1746.40& 945.41& 1828.36 \\
Max \# of Tokens        & 2031   & 1801  & 4046  \\
Min \# of Tokens        & 1519   & 717   & 817   \\
\bottomrule
\end{tabular}
}
\caption{Dataset statistics.}
\vspace{-1em}
\label{tab:dataset-stats}
\end{table}

\begin{itemize}[noitemsep,topsep=2pt,parsep=2pt,partopsep=2pt]
\item\textbf{ACL 60/60~\citep{salesky-etal-2023-evaluating}.}The first is ACL 60/60, which features 5 technical presentations from ACL 2022, each of which comes with an English talk and 10 corresponding translations to 10 languages. ACL 60/60 includes tags indicating NLP-related technical terms in the source and target talks, which we will use to evaluate terminology accuracy of translation methods. We consider 12 translation directions for the experiments: En $\Leftrightarrow$ Zh, De, Ja, Fr, Pt, and Ru.

\item\textbf{BWB~\citep{jiang-etal-2023-discourse}.} The BWB dataset is a large-scale document-level Chinese-English parallel dataset. It consists of 80 Chinese documents across multiple genres (sci-fi, romance, action, fantasy, comedy, etc.) and their corresponding English translations. Like ACL 60/60, this dataset also has tagged terminologies available. We consider 2 translation directions with the BWB benchmark: En $\Rightarrow$ Zh and Zh $\Rightarrow$ En.

\item\textbf{Guofeng Webnovel~\citep{wang2024findings}.} The Guofeng Webnovel includes high-quality documents and translations in English and Chinese. We use the TEST\_1 and TEST\_2 subsets of the benchmark, which accumulate 25 documents for evaluation. We consider 2 translation directions with the Guofeng Webnovel: En $\Rightarrow$ Zh and Zh $\Rightarrow$ En.
\end{itemize}

\vspace{-1.5mm}
\paragraph{Evaluation Metrics}
We opt for the classic document-level BLEU (d-BLEU) and document-level COMET (d-COMET) \cite{vernikos-etal-2022-embarrassingly}. Regarding the terminology-focused benchmarks (BWB and ACL 60/60), which have labeled terms in both source and reference documents, we add Terminology Accuracy as another evaluation metric, which computes the accuracy of translating terminologies when compared with terms in reference documents.

\subsection{Experiment Results}

\begin{table*}[t]
\centering
\resizebox{0.99\textwidth}{!}{%
\begin{tabular}{p{8cm}|ccc|ccc|cc}
\toprule
\multirow{2}{*}{\textbf{Method}} & \multicolumn{3}{c|}{\textbf{BWB}} & \multicolumn{3}{c|}{\textbf{ACL 60/60}}& \multicolumn{2}{c}{\textbf{GuoFeng}}\\
\cmidrule(lr){2-4}\cmidrule(lr){5-7}\cmidrule(lr){8-9}
 & \textbf{d-BLEU} & \textbf{d-COMET} & \textbf{Terminology Acc} & \textbf{d-BLEU} & \textbf{d-COMET} & \textbf{Terminology Acc}  & \textbf{d-BLEU} & \textbf{d-COMET} \\
\midrule
\multicolumn{9}{c}{\cellcolor{gray!30}\texttt{Qwen3-32B}}\\
\textsc{Sent. MT}         & 11.01 & 46.67 & 41.99 & 36.86 & 81.95 & 73.83  & 26.45 & 81.23\\
\textsc{1-Pass DocMT}   & 18.77 & 83.28 & 46.47 & 37.14 & 81.34 & 72.79  & 29.88 & 89.21\\
\textsc{TransAgent}                         & \textbf{19.06} & \textbf{84.73} & 53.21 & 37.13 & 81.82 & 74.69 & 31.57 & 90.32\\
\textsc{DeLTA}                              & 15.53 & 80.34 & 52.03 & \textbf{39.54} & 87.66 & 74.44 & 31.28 & 89.98\\
\hdashline
\ours (\textsc{$-$Rel})       & 19.27 & 83.78 & 51.05 & 37.27 & 83.87 & 74.55 & 30.72 & 89.72\\
\ours (\textsc{Fixed})                         & 14.92 & 79.96 & 60.20 & 37.55 & 88.94 & 78.06 & 32.54 & 90.38\\
\ours (\textsc{Seq})  & 14.87 & 79.36 & 52.47 & 37.09 & 88.98 & 74.41 & 31.87 & 90.04\\
\textbf{\textsc{TransGraph} (Ours)}         & 15.27 & 80.26 & \textbf{61.72} & 37.59 & \textbf{89.78} & \textbf{78.65} & \textbf{32.83} & \textbf{90.47}\\
\midrule
\multicolumn{9}{c}{\cellcolor{gray!30}\texttt{Qwen3-14B}}\\
\textsc{Sent. MT}       & 10.38 & 43.92 & 39.61 & 35.07 & 80.54 & 71.88 & 28.21 & 87.06\\
\textsc{1-Pass DocMT}  & 14.53 & 78.24 & 45.42 & 35.63 & 79.77 & 71.95 & 28.32 & 88.40\\
\textsc{TransAgent}                         & 14.44 & 79.12 & 52.13 & 35.52 & 80.21 & 73.47 & 28.86 & 89.25 \\
\textsc{DeLTA}                              & 17.73 & 81.92 & 50.72 & 36.61 & \textbf{88.92} & 73.82 & 28.98 & \textbf{88.58}\\
\hdashline
\ours (\textsc{$-$Rel})        & 16.18 & 79.55 & 51.92 & 36.37 & 81.67 & 73.65 & 29.07 & 88.02\\
\ours (\textsc{Fixed})                         & 17.69 & 82.17 & 59.15 & 37.84 & 85.48 & 77.04 & 29.73 & 88.10\\
\ours (\textsc{Seq})  & 17.54 & 81.71 & 51.27 & 37.59 & 85.31 & 73.49 & 28.96 & 88.05\\
\textbf{\textsc{TransGraph} (Ours)}         & \textbf{17.94} & \textbf{82.61} & \textbf{60.68} & \textbf{38.09} & 86.11 & \textbf{77.61} & \textbf{30.08} & 88.19 \\
\midrule
\multicolumn{9}{c}{\cellcolor{gray!30}\texttt{Qwen3-8B}}\\
\textsc{Sent. MT}          &  9.26 & 41.12 & 35.94 & 33.24 & 78.87 & 70.63 & 22.04 & 82.11 \\
\textsc{1-Pass DocMT}   & 16.22 & 78.41 & 43.11 & 33.98 & 78.56 & 69.92 & 21.62 & 81.44\\
\textsc{TransAgent}                         & 13.14 & 75.16 & 50.24 & 33.72 & 79.23 & 72.02 & 23.46 & 81.23\\
\textsc{DeLTA}                              & 12.93 & 76.62 & 48.41 & 35.04 & \textbf{87.71} & 72.69 & 26.22 & \textbf{81.80}\\
\hdashline
\ours (\textsc{$-$Rel})        & 16.40 & 76.10 & 47.63 & 34.62 & 80.52 & 71.78 & 20.22 & 81.61 \\
\ours (\textsc{Fixed})                        & 16.77 & 78.68 & 56.68 & 35.90 & 84.43 & 75.51 & 25.92 & 79.18\\
\ours (\textsc{Seq})  & 16.43 & 78.21 & 49.18 & 35.61 & 84.28 & 72.20 & 25.58 & 79.11\\
\textbf{\textsc{TransGraph} (Ours)}         & \textbf{16.83} & \textbf{79.11} & \textbf{58.19} & \textbf{36.11} & 85.08 & \textbf{76.13} & \textbf{26.24} & 79.29 \\
\midrule
\multicolumn{9}{c}{\cellcolor{gray!30}\texttt{Llama3.1-Instruct-8B}}\\
\textsc{Sent. MT}               & 12.79 & 73.57 & 36.22 & 19.30 & 0.62 & 57.15 & 22.10 & 81.41 \\
\textsc{1-Pass DocMT}           & 13.30 & 73.31 & 43.58 & 19.47 & 0.77 & 62.33  & 23.54 & 82.19\\
\textsc{TransAgent}                         & 13.78 & 74.70 & 50.09 & 19.98 & 0.76 & 62.98 & \textbf{23.85} & 82.45\\
\textsc{DeLTA}                              & 12.79 & \textbf{75.78} & 48.33 & 22.33 & \textbf{0.78} & 67.68 & 22.14 & 81.32\\
\hdashline
\ours (\textsc{$-$Rel})        & 13.68 & 73.32 & 48.13 & 20.38 & 0.77 & 67.05 & 23.74 & 83.05 \\
\ours (\textsc{Fixed})                         & 14.45 & 73.34 & 57.22 & 22.21 & 0.78 & 76.13 & 22.95 & 83.18  \\
\ours (\textsc{Seq})  & 14.18 & 72.44 & 49.06 & 21.91 & 0.77 & 65.18 & 22.87 & 83.03\\
\textbf{\textsc{TransGraph} (Ours)}         & \textbf{14.58} & 73.34 & \textbf{58.74} & \textbf{22.51} & \textbf{0.78} & \textbf{77.83} & 23.07 & \textbf{83.26}\\
\midrule
\multicolumn{9}{c}{\cellcolor{gray!30}\texttt{Ministral-Instruct-8B}}\\
\textsc{Sent. MT}                 &  9.51 & 75.40 & 34.87 & 36.08 & 0.79 & 67.67 & 21.99 & 79.22\\
\textsc{1-Pass DocMT}           &  9.62 & 71.50 & 42.15 & 35.52 & 0.75 & 62.55 & 21.81 & 80.17\\
\textsc{TransAgent}                                  &  9.83 & 75.35 & 49.18 & 41.40 & 0.87 & 77.43 & 22.65 & 80.34\\
\textsc{DeLTA}                                       & \textbf{15.39} & 74.85 & 47.60 & 40.13 & 0.74 & 67.70 & 22.15 & 81.21\\
\hdashline
\ours (\textsc{$-$Rel})        & 10.99 & 75.76 & 46.62 & 39.22 & 0.79 & 68.91 & 22.37 & 80.92\\
\ours (\textsc{Fixed})                         & 13.92 & 76.27 & 55.57 & 46.61 & 0.87 & 80.39 & 22.82 & 81.28\\
\ours (\textsc{Seq})  & 14.01 & 75.46 & 48.24 & 47.24 & 0.87 & 72.41 & 22.71 & 81.30\\
\textbf{\textsc{TransGraph} (Ours)}         & 14.41 & \textbf{76.36} & \textbf{57.06} & 47.84 & \textbf{0.88} & \textbf{82.57} & \textbf{22.99} & \textbf{81.37} \\
\bottomrule
\end{tabular}
}
\caption{Results of different document-level MT methods on terminology-focused benchmarks (BWB, ACL 60/60 benchmark) and regular benchmark (GuoFeng TEST\_1 and TEST\_2).}
\label{tab-exp-acl6060}
\vspace{-1em}
\end{table*}

Across all backbones and datasets, \textsc{TransGraph} delivers consistent gains over both sentence-level and single-pass document baselines, and is competitive with, often superior to, recent agentic systems.

\vspace{-1.5mm}
\paragraph{Regular Document Translation.}
On the combined GuoFeng test sets (\Cref{tab-exp-acl6060}), \textsc{TransGraph} improves d-BLEU and d-COMET for most LLM backbones. With \texttt{Qwen3-32B}, \textsc{TransGraph} attains the best overall score,
exceeding the sentence baseline by +6.38 BLEU and +9.24 COMET, and edging out \textsc{TransAgent} and \textsc{DeLTA}. Trends hold for \texttt{Llama3.1-8B} and \texttt{Ministral-8B} backbones, where \textsc{TransGraph} yields the highest d-COMET (83.26 and 81.37, respectively). One exception appears with \texttt{Qwen3-8B}, where \textsc{TransGraph} tops d-BLEU (26.24) but trails \textsc{DeLTA} on d-COMET (79.29 vs.\ 81.80), suggesting that strong local faithfulness and style modelling remain challenging for the smallest backbone when only shallow context is available. Overall, the pattern indicates that graph-conditioned retrieval provides stable quality benefits without expensive and costly multi-agent orchestration.

\vspace{-1.5mm}
\paragraph{Terminology-focused Document Translation.}
Terminology-focused translation is where \textsc{TransGraph} is most distinctive. As shown in \Cref{tab-exp-acl6060}, on the BWB benchmark, with the \texttt{Qwen3-32B} backbone, \textsc{TransGraph} exhibits a termninology accuracy of 61.72, outperforming both \textsc{TransAgent} and \textsc{DeLTA}, as well as the other two baselines. \textsc{TransGraph} also showed outstanding performance in terms of terminology with other LLM backbones, including both medium and lightweight models. On the ACL60/60 benchmark, \textsc{TransGraph} also has better accuracy across different backbones (e.g., \texttt{Qwen3-32B}: 78.65 vs.\ 74–75 with agentic baselines). These improvements arrive without sacrificing d-BLEU/d-COMET; for instance, on ACL60/60 with \texttt{Qwen3-32B}, d-COMET of \textsc{TransGraph} reaches 89.78 while terminology accuracy remains in first place, indicating that explicit discourse guided translation helps preserve specialized terms without degrading general adequacy/fluency.

\subsection{Discussions}

In this section, we conduct various intrinsic experiments of \textsc{TransGraph} to justify our design choices, as well as perform a cost analysis and cohesion analysis of \textsc{TransGraph} and other document translation methods.

\subsubsection{Chunking Analysis}

In this section, we measure the document chunking performance of LLMs by performing an evaluation on the ACL 60/60 benchmark. We first perform a manual annotation to label the position of each chunk for the 5 English documents, then calculate the overlapping rate of chunks split by LLMs in different languages. Specifically, we give the models the documents in different languages available in the ACL 60/60 test set (En, Zh, De, Ja, Pt, and Ru) and calculate the overlapping rate of generated chunks w.r.t. our manual annotation by (i) iterating through each pair of chunks and (ii) computing the ratio of overlapping sentences over the maximum number of sentences between two chunks. \Cref{tab:overlap-chunk} shows the results. As expected, chunking performance decreases gradually with smaller model sizes, reflecting the translation performance shown in \Cref{tab-exp-acl6060}. En and Zh have the best chunking quality across LLMs, as Qwen-based models are trained heavily on English and Chinese.

\begin{table}[t]
\centering
\resizebox{0.99\columnwidth}{!}{
\begin{tabular}{p{4cm}ccc}
\toprule
\textbf{Models} & \textbf{Qwen3-8B} & \textbf{Qwen3-14B} & \textbf{Qwen3-32B} \\
\midrule
Annotated $\Leftrightarrow$ En & 80.23 & 87.73 & 90.96 \\
Annotated $\Leftrightarrow$ De & 60.62 & 72.11 & 81.93 \\
Annotated $\Leftrightarrow$ Zh & 79.37 & 84.60 & 91.72 \\
Annotated $\Leftrightarrow$ Ru & 64.15 & 71.94 & 76.87 \\
Annotated $\Leftrightarrow$ Pt & 74.63 & 83.12 & 88.13 \\
Annotated $\Leftrightarrow$ Ja & 58.28 & 63.19 & 66.46 \\
Annotated $\Leftrightarrow$ Fr & 65.91 & 74.32 & 81.02 \\
\bottomrule
\end{tabular}
}
\caption{Average overlaping rate across language pairs and model sizes.}
\label{tab:overlap-chunk}
\vspace{-1em}
\end{table}

\subsubsection{Graph Analysis}

\begin{table}[htbp]
\centering
\footnotesize
\resizebox{0.99\columnwidth}{!}{
\begin{tabular}{p{4cm}ccc}
\toprule
\textbf{Model} & \textbf{Qwen3-32B} & \textbf{Qwen3-14B} & \textbf{Qwen3-8B} \\
\midrule
\multicolumn{4}{l}{\emph{Graph Accuracy}} \\
\midrule
En-Acc & 94.54 & 90.45 & 87.27 \\
De-Acc & 89.09 & 88.18 & 77.72 \\
Zh-Acc & 90.45 & 90.90 & 80.45 \\
\addlinespace[2pt]
\midrule
\multicolumn{4}{l}{\emph{Graph Consistency}} \\
\midrule
En \& De & 76.81 & 75.00 & 70.45 \\
En \& Zh & 87.15 & 86.85 & 79.18 \\
\bottomrule
\end{tabular}
}
\caption{Graph Accuracy and Consistency of different LLM backbones.}
\label{tab:graph-analysis}
\vspace{-1em}
\end{table}
\paragraph{Relation identification accuracy.} We first evaluate the performance of LLMs in determining the relations between chunks to construct the discourse graph. From the ACL60/60 benchmark, we manually annotated 20 samples for each relation type, resulting in a test set of 220 samples. Each sample comes with a pair of chunks in English, Chinese, and German, along with their labeled relation type. \Cref{tab:graph-analysis} shows the evaluation results. \texttt{Qwen3-32B} achieves 94.54\% on English, 89.09\% on German, and 90.45\% on Chinese accuracy. Accuracy degrades gradually  with smaller models (\texttt{Qwen3-14B} around 88–91\%; \texttt{Qwen3-8B} around 78–87\%), mirroring translation-quality trends. This sensitivity analysis is important because false relation labels can introduce misleading context; empirically, however, \textsc{TransGraph}'s reliance on a bounded in-neighborhood and label-aware prompts appears robust to sporadic mislabeling.
\vspace{-1.5mm}
\paragraph{Graph–graph consistency across languages.} 
From the ACL60/60 benchmark, we compare discourse graphs induced from English source documents to graphs induced from Chinese and German references. Graphs are constructed by going through Stage 1 of \textsc{TransGraph}, and Graph Consistency is measured by calculating the overlapping rate of two sets of relations deduced from a pair of discourse graphs in different languages. \texttt{Qwen3-32B} attains 76.81\% (En \& De) and 87.15\% (En \& Zh) consistency, while \texttt{Qwen3-14B} is close (75.00/86.85\%) and \texttt{Qwen3-8B} remains reasonable (70.45/79.18\%). The higher En \& Zh consistency likely reflects the raw performance of \texttt{Qwen3}, as this model family is trained primarily on English and Chinese. These findings support our design choices of \textsc{TransGraph} - when the LLM is conditioned on explicit discourse roles (Background→Core, Core→Detail, etc.), rhetorical structure is more faithfully preserved cross-lingually.

\subsubsection{Design Choices Justification}


We justify the design choice of \textsc{TransGraph} by comparing it with the following baselines: (i) \ours (\textsc{Fixed}), (ii) \ours (\textsc{$-$Rel}), and (iii) \ours (\textsc{Seq}).
\vspace{-1.5mm}
\paragraph{Importance of Coherent Chunking.} \Cref{tab-exp-acl6060} shows that naively performing document chunking with uniform-length chunks (\ours (\textsc{Fixed})) exhibits worse performance compared to \textsc{TransGraph} on all benchmarks. Although the evaluation results of this baseline still consistently outperform other baselines, especially on Terminology Accuracy, it is not optimal since the chunks are not split based on content, leading to unnecessary sentences in chunks, which confuses the LLMs in identifying the discourse relations.
\vspace{-1.5mm}
\paragraph{Importance of Discourse Relations.} We also experimented with removing the discourse graph and just providing LLMs with the 5 previous chunks' content when translating a target chunk (\ours (\textsc{$-$Rel})). As shown in \Cref{tab-exp-acl6060}, performance is much worse than \textsc{TransGraph}, and it is only slightly better than \textsc{1-Pass DocMT} baseline. This proves that discourse relations are important in maintaining translation quality and terminology accuracy.
\vspace{-1.5mm}
\paragraph{Importance of the Graph.} Finally, in order to measure the effectiveness of the graph design in document translation, we compare \textsc{TransGraph} with \ours (\textsc{Seq}) baseline . As shown in \Cref{tab-exp-acl6060}, while the d-BLEU and d-COMET scores of this baseline are slightly lower than the original \textsc{TransGraph}, the Terminology Accuracy is significantly worse. This is because when we only condition the translation of a chunk on the nearest chunks, the LLM does not have access to faraway chunks that may directly contribute to the translation of specific terms.

\begin{figure}[t]    
\includegraphics[width=0.99\columnwidth]{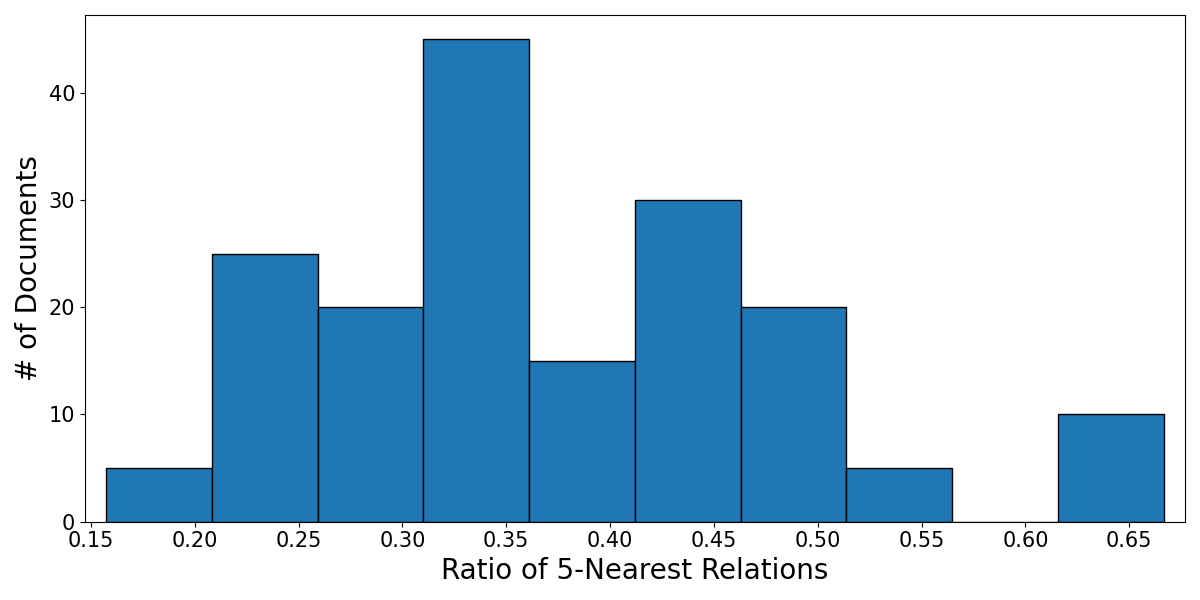}
\vspace{-0.5em}
\caption{Distribution of ratios of 5-nearest relations out of the total number of relations. Distribution is calculated on the documents of BWB and ACL 60/60 benchmarks.}
\label{fig:ratio}
\vspace{-1em}
\end{figure}

\Cref{fig:ratio} illustrates the ratio distribution of the number of existing relations among 5-nearest chunks compared to the total number of relations in the BWB and ACL 60/60 benchmarks. Most of the documents have low ratios around 35\%, which means that if we only consider 5-nearest chunks in the graph of \textsc{TransGraph}, we are missing out 65\% of the relations, resulting in poorer performance and terminology consistency.

\subsubsection{Cost Analysis}
\begin{table*}[t]
\centering
\resizebox{\textwidth}{!}{%
\begin{tabular}{@{}l|cccccc|cccccc@{}}
\toprule
\textbf{Dims.} & \multicolumn{6}{c|}{\textbf{Coreference}} & \multicolumn{6}{c}{\textbf{Conjunction}} \\ \midrule
\textbf{\begin{tabular}[c]{@{}l@{}}Lang. \\ (En-XX)\end{tabular}} & \textbf{Sent-level} & \textbf{\begin{tabular}[c]{@{}c@{}}Single-Pass \\ Doc-level\end{tabular}} & \multicolumn{1}{l}{\textbf{\textsc{TransAgent}}} & \multicolumn{1}{l}{\textbf{\textsc{DeLTA}}} & \multicolumn{1}{c|}{\textbf{\begin{tabular}[c]{@{}c@{}}\textsc{TransGraph}\\ (ours)\end{tabular}}} & \textbf{Reference} & \textbf{Sent-level} & \textbf{\begin{tabular}[c]{@{}c@{}}Single-Pass \\ Doc-level\end{tabular}} & \multicolumn{1}{l}{\textbf{\textsc{TransAgent}}} & \multicolumn{1}{l}{\textbf{\textsc{DeLTA}}} & \multicolumn{1}{c|}{\textbf{\begin{tabular}[c]{@{}c@{}}\textsc{TransGraph}\\ (ours)\end{tabular}}} & \textbf{Reference} \\ \midrule
\textbf{De} & 80.33 & \textbf{99.35} & 96.03 & 95.80 & \multicolumn{1}{c|}{97.37} & 97.68 & 80.41 & 75.18 & 82.10 & \textbf{82.95} & \multicolumn{1}{c|}{82.21} & 87.02 \\
\textbf{Fr} & 78.85 & 98.03 & 96.81 & 98.69 & \multicolumn{1}{c|}{\textbf{98.06}} & 97.56 & 75.87 & 76.76 & 76.37 & \textbf{84.04} & \multicolumn{1}{c|}{81.32} & 95.52 \\
\textbf{Ja} & 74.88 & 98.73 & 97.79 & 98.08 & \multicolumn{1}{c|}{\textbf{100.00}} & 100.00 & 63.69 & 63.27 & 77.58 & 76.32 & \multicolumn{1}{c|}{\textbf{78.99}} & 84.28 \\
\textbf{Pt} & 74.08 & 97.22 & 96.53 & 97.89 & \multicolumn{1}{c|}{\textbf{98.97}} & \textbf{97.57} & 73.51 & 83.21 & 85.84 & \textbf{89.61} & \multicolumn{1}{c|}{\textbf{85.77}} & 92.91 \\
\textbf{Ru} & 79.57 & 98.11 & 97.99 & 97.93 & \multicolumn{1}{c|}{\textbf{98.74}} & \textbf{97.91} & 75.35 & 75.02 & 80.74 & 83.50 & \multicolumn{1}{c|}{\textbf{83.92}} & 77.43 \\
\textbf{Zh} & 82.65 & 98.52 & \textbf{100.00} & 99.57 & \multicolumn{1}{c|}{\textbf{100.00}} & 99.56 & 67.04 & 72.62 & 80.54 & 77.62 & \multicolumn{1}{c|}{\textbf{81.51}} & 80.64 \\ \midrule
\textbf{Avg.} & 78.39 & 98.33 & 97.52 & 97.99 & \multicolumn{1}{c|}{\textbf{98.86}} & 98.38 & 72.65 & 74.34 & 80.53 & \textbf{82.34} & \multicolumn{1}{c|}{82.29} & 86.30 \\ \bottomrule
\end{tabular}%
}
\caption{Cohesion evaluation on the ACL 60/60 dataset, scores represent translation accuracy on coreference and conjunction with a range between 0 and 100. Best-performed methods within each language pair are \textbf{bolded} (dataset reference is excluded).}
\label{tab:cohesion_evaluation}
\vspace{-1em}
\end{table*}
\begin{table}[t]
\centering
\footnotesize
\resizebox{0.49\textwidth}{!}{
\begin{tabular}{p{3cm}rrrr}
\toprule
\textbf{} & \textbf{Avg Input} & \textbf{Avg Output} & \textbf{Avg LLM} & \textbf{Avg Total} \\
\textbf{Method} & \textbf{Tokens} & \textbf{Tokens} & \textbf{Calls} & \textbf{Tokens} \\
\midrule
\textsc{Sent. MT}         & 45   & 36   & 32  & 2,592   \\
\textsc{1-Pass DocMT}   & 1,010 & 815  & 1   & 1,825   \\
\textsc{TransAgent}                         & 932  & 251  & 88  & 104,104 \\
\textsc{DeLTA}                              & 621  & 24   & 266 & 171,304 \\
\textbf{\textsc{TransGraph} (Ours)}         & 1,351 & 98   & 42  & 59,368  \\
\bottomrule
\end{tabular}
}
\caption{Cost analysis of different DocMT methods.}
\label{tab-speed}
\vspace{-1em}
\end{table}

We analyze the cost of different document translation methods to highlight the efficiency of \textsc{TransGraph} (\Cref{tab-speed}). We report the average number of input \& output tokens, as well as the average number of LLM calls and total tokens when running on the three document translation benchmarks. \textsc{TransAgent} and \textsc{DeLTA} incur large budgets due to back-and-forth multi-agent exchanges and sentence-granular passes - average total tokens $\approx$ 104k and 171k per document, respectively. By contrast, \textsc{TransGraph} averages $\approx$ 59k tokens with 42 calls on average - substantially below agentic baselines, while maintaining higher accuracy on terminology and stronger BLEU/COMET. As expected, single-pass document translation is cheapest ($\approx$1.8k tokens, 1 call) but underperforms markedly on document-level metrics and terminology. Sentence-level decoding is also cheap per call but accumulates calls (avg 32) and suffers from maintaining cohesion and terminology accuracy. In short, \textsc{TransGraph} occupies a favorable quality–cost regime - one-order-of-magnitude fewer tokens than multi-agent frameworks, yet with stronger document translation performance.

\subsubsection{Cohesion Analysis}

We further analyzed cohesion in different translation methods on the ACL 60/60 benchmark. Unlike BLEU and COMET, evaluating discourse phenomena like coreference and conjunction requires deep semantic understanding. While existing methods~\citep{tan2022discoursecohesionevaluationdocumentlevel,jiang2023discoursecentricevaluationmachine} assess whether translations preserve entity dependencies and logical relationships, their complex annotation pipelines limit practical use. We simplify this process using LLM-as-a-Judge: an LLM first annotates pronouns with their referents and conjunctions with logical relationships in the source text, then checks translation accuracy in the target text using this annotation as grounding. We use \texttt{Gemini-2.5-Flash} for its strong capabilities and long context length (detailed prompt in~\Cref{app:prompt_templates}). As shown in~\Cref{tab:cohesion_evaluation}, for coreference, doc-level methods significantly outperform sentence-level translation, with \textsc{TransGraph} surpassing baselines across multiple languages and achieving perfect scores in Japanese and Chinese. For conjunction, \textsc{TransGraph} achieves performance comparable to \textsc{DeLTA}, with both agentic and graph-based methods expressing logical relationships more accurately than simple sentence/doc-level approaches. Interestingly, translation systems sometimes score higher than the reference because ACL 60/60 contains spoken language where references retain colloquial errors that reduce readability, while translation systems produce more polished, written-style output, resulting in higher scores.
\section{Related Works}
\paragraph{DocMT and context integration}
Early DocMT incorporated sentence–external context via caches, hierarchical attention, and memory–augmented Transformers \citep{tu2018cache,miculicich2018document,voita2018context,kim2019when,zhang2022rethinking}. \cite{maruf2019survey} presents a comprehensive survey of DocMT.  Beyond generic context, graph and discourse–aware methods encode structure explicitly: document graphs and coreference–aware encoders improve coherence \citep{ohtani2019coref,docgraph2021}. Large–scale document corpora and para–document resources continue to expand evaluation coverage \citep{paraDocMT2024}. 
\vspace{-1.5mm}
\paragraph{LLM-based DocMT}
LLMs have recently demonstrated strong performance on MT tasks~\citep{pang2025salute}. Efforts to extend their capabilities to DocMT include fine-tuning on high-quality parallel data~\citep{wu2024adapting,ramos2025multilingual} and context-aware promoting~\citep{wang-etal-2024-benchmarking,cui-etal-2024-efficiently,wang-etal-2023-document-level}. More recently, growing attention has turned toward multi–agent DocMT pipelines which coordinates roles
and memories 
to sustain translation consistency \citep{transagents_demo,transagent,delta_arxiv}. While effective, these designs can be token–intensive. 
Concurrently, \citet{graft2025} also employs LLMs to build document-level relation graphs, but unlike our approach, they model only the existence of links between chunks without explicitly defining relation types, and rely on external memory modules for context summarisation.
In contrast, our work introduces discord-graph guided approach, offering structured, compact context that enhances cohesion and terminology accuracy with minimal overhead, aligning naturally with discourse theory~\citep{mann1988rst,taboada2006rst}.
\section{Conclusions}

We presented \textsc{TransGraph}, a discourse–guided DocMT framework that conditions chunk translations on compact, relation–labeled graph context. Experiments on ACL 60/60, BWB, and GuoFeng show consistent improvements in document–level BLEU/COMET and substantial gains in terminology accuracy relative to sentence–level, single–pass, and agentic baselines, while reducing total tokens and calls. LLM–assisted cohesion evaluation confirms benefits for coreference and conjunction handling. Our findings suggest that modest, well–structured discourse retrieval is a robust lever for document translation with LLMs.

\section*{Limitations}

First, discourse graphs rely on LLM relation labeling. Although our analysis shows high accuracy and reasonable cross–lingual consistency, residual mislabels may inject noisy context. Second, while graph–retrieved context reduces token budgets relative to agentic pipelines, scalability still depends on chunking granularity and windowing. Very long documents with dense relation structure increase retrieval and prompt assembly costs. Moreover, LLMs remain sensitive to context placement and length \citep{dai2019transformerxl,beltagy2020longformer,liu2023lost}. Future works could combine retrieval–augmented compression and learned relation selection to further improve robustness and efficiency.

\bibliography{custom}

@misc{tan2022discoursecohesionevaluationdocumentlevel,
      title={Discourse Cohesion Evaluation for Document-Level Neural Machine Translation}, 
      author={Xin Tan and Longyin Zhang and Guodong Zhou},
      year={2022},
      eprint={2208.09118},
      archivePrefix={arXiv},
      primaryClass={cs.CL},
      url={https://arxiv.org/abs/2208.09118}, 
}

@misc{jiang2023discoursecentricevaluationmachine,
      title={Discourse Centric Evaluation of Machine Translation with a Densely Annotated Parallel Corpus}, 
      author={Yuchen Eleanor Jiang and Tianyu Liu and Shuming Ma and Dongdong Zhang and Mrinmaya Sachan and Ryan Cotterell},
      year={2023},
      eprint={2305.11142},
      archivePrefix={arXiv},
      primaryClass={cs.CL},
      url={https://arxiv.org/abs/2305.11142}, 
}

@article{maruf2019survey,
  title = {A Survey on Document-level Neural Machine Translation: Methods and Evaluation},
  author = {Maruf, Sameen and Saleh, Fahimeh and Haffari, Gholamreza},
  journal = {arXiv},
  year = {2019},
  eprint = {1912.08494},
  url = {https://arxiv.org/abs/1912.08494}
}

@inproceedings{tu2018cache,
  title = {Learning to Remember Translation History with a Continuous Cache},
  author = {Tu, Zhaopeng and Liu, Yang and Liu, Shuming and Zhou, Maosong},
  booktitle = {Transactions of the Association for Computational Linguistics},
  volume = {6},
  pages = {407--420},
  year = {2018},
  url = {https://aclanthology.org/Q18-1029/}
}

@inproceedings{miculicich2018document,
  title = {Document-Level Neural Machine Translation with Hierarchical Attention Networks},
  author = {Miculicich, Lesly and Ram, Dhananjay and Pappas, Nikolaos and Henderson, James},
  booktitle = {Proceedings of EMNLP},
  year = {2018},
  pages = {2947--2954},
  url = {https://aclanthology.org/D18-1325/}
}

@inproceedings{voita2018context,
  title = {Context-Aware Neural Machine Translation Learns Anaphora Resolution},
  author = {Voita, Elena and Serdyukov, Pavel and Sennrich, Rico and Titov, Ivan},
  booktitle = {Proceedings of ACL},
  year = {2018},
  pages = {1264--1274},
  url = {https://aclanthology.org/P18-1117/}
}

@inproceedings{kim2019when,
  title = {When and Why is Document-level Context Useful in Neural Machine Translation?},
  author = {Kim, Yunsu and Happe, Kilian and Petrushkov, Pavel and van Genabith, Josef and Ney, Hermann},
  booktitle = {Proceedings of the Fourth Workshop on Discourse in Machine Translation},
  year = {2019},
  pages = {24--34},
  url = {https://aclanthology.org/D19-6503/}
}

@inproceedings{zhang2022rethinking,
  title = {Rethinking Document-level Neural Machine Translation},
  author = {Zhang, Shaolei and Feng, Yang and others},
  booktitle = {Findings of ACL},
  year = {2022},
  pages = {3537--3548},
  url = {https://aclanthology.org/2022.findings-acl.279/}
}

@inproceedings{dai2019transformerxl,
  title = {Transformer-{XL}: Attentive Language Models Beyond a Fixed-Length Context},
  author = {Dai, Zihang and Yang, Zhilin and Yang, Yiming and Carbonell, Jaime and Le, Quoc V. and Salakhutdinov, Ruslan},
  booktitle = {Proceedings of ACL},
  year = {2019},
  pages = {2978--2988},
  url = {https://aclanthology.org/P19-1285/}
}

@article{beltagy2020longformer,
  title = {Longformer: The Long-Document Transformer},
  author = {Beltagy, Iz and Peters, Matthew E. and Cohan, Arman},
  journal = {arXiv},
  year = {2020},
  eprint = {2004.05150},
  url = {https://arxiv.org/abs/2004.05150}
}

@inproceedings{jiang2023discourse,
  title = {Discourse-Centric Evaluation of Document-level Machine Translation with a New Densely Annotated Parallel Corpus of Novels},
  author = {Jiang, Yuchen Eleanor and Liu, Tianyu and Ma, Shuming and Zhang, Dongdong and Sachan, Mrinmaya and Cotterell, Ryan},
  booktitle = {Proceedings of ACL (Long Papers)},
  year = {2023},
  pages = {7853--7872},
  address = {Toronto, Canada},
  url = {https://aclanthology.org/2023.acl-long.435/}
}

@inproceedings{transagents_demo,
  title = {TransAgents: Build Your Translation Company with Language Agents},
  author = {Wu, Minghao and Xu, Jiahao and Yuan, Yulin and Haffari, Gholamreza and Wang, Longyue and Luo, Weihua and Zhang, Kaifu},
  booktitle = {Proceedings of EMNLP 2024: System Demonstrations},
  year = {2024},
  url = {https://aclanthology.org/2024.emnlp-demo.14/}
}

@inproceedings{vernikos-etal-2022-embarrassingly,
    title = "Embarrassingly Easy Document-Level {MT} Metrics: How to Convert Any Pretrained Metric into a Document-Level Metric",
    author = "Vernikos, Giorgos  and
      Thompson, Brian  and
      Mathur, Prashant  and
      Federico, Marcello",
    editor = {Koehn, Philipp  and
      Barrault, Lo{\"i}c  and
      Bojar, Ond{\v{r}}ej  and
      Bougares, Fethi  and
      Chatterjee, Rajen  and
      Costa-juss{\`a}, Marta R.  and
      Federmann, Christian  and
      Fishel, Mark  and
      Fraser, Alexander  and
      Freitag, Markus  and
      Graham, Yvette  and
      Grundkiewicz, Roman  and
      Guzman, Paco  and
      Haddow, Barry  and
      Huck, Matthias  and
      Jimeno Yepes, Antonio  and
      Kocmi, Tom  and
      Martins, Andr{\'e}  and
      Morishita, Makoto  and
      Monz, Christof  and
      Nagata, Masaaki  and
      Nakazawa, Toshiaki  and
      Negri, Matteo  and
      N{\'e}v{\'e}ol, Aur{\'e}lie  and
      Neves, Mariana  and
      Popel, Martin  and
      Turchi, Marco  and
      Zampieri, Marcos},
    booktitle = "Proceedings of the Seventh Conference on Machine Translation (WMT)",
    month = dec,
    year = "2022",
    address = "Abu Dhabi, United Arab Emirates (Hybrid)",
    publisher = "Association for Computational Linguistics",
    url = "https://aclanthology.org/2022.wmt-1.6/",
    pages = "118--128"
}

@inproceedings{delta_arxiv,
title={Del{TA}: An Online Document-Level Translation Agent Based on Multi-Level Memory},
author={Yutong Wang and Jiali Zeng and Xuebo Liu and Derek F. Wong and Fandong Meng and Jie Zhou and Min Zhang},
booktitle={The Thirteenth International Conference on Learning Representations},
year={2025},
url={https://openreview.net/forum?id=hoYFLRNbhc}
}

@inproceedings{codonmt2022,
  title = {{CoDoNMT}: Modeling Cohesion Devices for Document-Level Neural Machine Translation},
  author = {Qin, Libo and Ji, Zecheng and Wang, Wanqi and Che, Wanxiang and Liu, Ting},
  booktitle = {Proceedings of COLING},
  year = {2022},
  pages = {5232--5245},
  url = {https://aclanthology.org/2022.coling-1.462/}
}

@article{mann1988rst,
  title = {Rhetorical Structure Theory: Toward a Functional Theory of Text Organization},
  author = {Mann, William C. and Thompson, Sandra A.},
  journal = {Text \& Talk},
  volume = {8},
  number = {3},
  pages = {243--281},
  year = {1988},
  doi = {10.1515/text.1.1988.8.3.243},
  url = {https://www.degruyterbrill.com/document/doi/10.1515/text.1.1988.8.3.243/html}
}

@article{taboada2006rst,
  title = {Rhetorical Structure Theory: Looking Back and Moving Ahead},
  author = {Taboada, Maite and Mann, William C.},
  journal = {Discourse Studies},
  volume = {8},
  number = {3},
  pages = {423--459},
  year = {2006},
  url = {https://www.sfu.ca/~mtaboada/docs/publications/Taboada_Mann_RST_Part1.pdf}
}

@inproceedings{ohtani2019coref,
    title = "Context-aware Neural Machine Translation with Coreference Information",
    author = "Ohtani, Takumi  and
      Kamigaito, Hidetaka  and
      Nagata, Masaaki  and
      Okumura, Manabu",
    editor = "Popescu-Belis, Andrei  and
      Lo{\'a}iciga, Sharid  and
      Hardmeier, Christian  and
      Xiong, Deyi",
    booktitle = "Proceedings of the Fourth Workshop on Discourse in Machine Translation (DiscoMT 2019)",
    month = nov,
    year = "2019",
    address = "Hong Kong, China",
    publisher = "Association for Computational Linguistics",
    url = "https://aclanthology.org/D19-6505/",
    doi = "10.18653/v1/D19-6505",
    pages = "45--50"
}

@article{docgraph2021,
      title={Document Graph for Neural Machine Translation}, 
      author={Mingzhou Xu and Liangyou Li and Derek. F. Wong and Qun Liu and Lidia S. Chao},
      year={2021},
      eprint={2012.03477},
      archivePrefix={arXiv},
      primaryClass={cs.CL},
      url={https://arxiv.org/abs/2012.03477}, 
}

@article{paraDocMT2024,
      title={Exploring Paracrawl for Document-level Neural Machine Translation}, 
      author={Yusser Al Ghussin and Jingyi Zhang and Josef van Genabith},
      year={2023},
      eprint={2304.10216},
      archivePrefix={arXiv},
      primaryClass={cs.CL},
      url={https://arxiv.org/abs/2304.10216}, 
}

@article{graft2025,
      title={GRAFT: A Graph-based Flow-aware Agentic Framework for Document-level Machine Translation}, 
      author={Himanshu Dutta and Sunny Manchanda and Prakhar Bapat and Meva Ram Gurjar and Pushpak Bhattacharyya},
      year={2025},
      eprint={2507.03311},
      archivePrefix={arXiv},
      primaryClass={cs.CL},
      url={https://arxiv.org/abs/2507.03311}, 
}

@misc{liu2023lost,
      title={Lost in the Middle: How Language Models Use Long Contexts}, 
      author={Nelson F. Liu and Kevin Lin and John Hewitt and Ashwin Paranjape and Michele Bevilacqua and Fabio Petroni and Percy Liang},
      year={2023},
      eprint={2307.03172},
      archivePrefix={arXiv},
      primaryClass={cs.CL},
      url={https://arxiv.org/abs/2307.03172}, 
}

@article{qwen3,
    title={Qwen3 Technical Report}, 
    author={An Yang and Anfeng Li and Baosong Yang and Beichen Zhang and Binyuan Hui and Bo Zheng and Bowen Yu and Chang Gao and Chengen Huang and Chenxu Lv and others},
    journal = {arXiv preprint arXiv:2505.09388},
    year={2025}
}

@article{grattafiori2024llama,
  title={The llama 3 herd of models},
  author={Grattafiori, Aaron and Dubey, Abhimanyu and Jauhri, Abhinav and Pandey, Abhinav and Kadian, Abhishek and Al-Dahle, Ahmad and Letman, Aiesha and Mathur, Akhil and Schelten, Alan and Vaughan, Alex and others},
  journal={arXiv preprint arXiv:2407.21783},
  year={2024}
}

@article{transagent,
    author = {Wu, Minghao and Xu, Jiahao and Yuan, Yulin and Haffari, Gholamreza and Wan, Longyue and Luo, Weihua and Zhang, Kaifu},
    title = {(Perhaps) Beyond Human Translation: Harnessing Multi-Agent Collaboration for Translating Ultra-Long Literary Texts},
    journal = {Transactions of the Association for Computational Linguistics},
    volume = {13},
    pages = {901-922},
    year = {2025},
    month = {07},
    issn = {2307-387X},
    doi = {10.1162/TACL.a.25},
    url = {https://doi.org/10.1162/TACL.a.25},
    eprint = {https://direct.mit.edu/tacl/article-pdf/doi/10.1162/TACL.a.25/2540062/tacl.a.25.pdf},
}

@inproceedings{salesky-etal-2023-evaluating,
    title = "Evaluating Multilingual Speech Translation under Realistic Conditions with Resegmentation and Terminology",
    author = "Salesky, Elizabeth  and
      Darwish, Kareem  and
      Al-Badrashiny, Mohamed  and
      Diab, Mona  and
      Niehues, Jan",
    booktitle = "Proceedings of the 20th International Conference on Spoken Language Translation (IWSLT 2023)",
    month = jul,
    year = "2023",
    address = "Toronto, Canada",
    publisher = "Association for Computational Linguistics",
    url = "https://aclanthology.org/2023.iwslt-1.2/",
    doi = "10.18653/v1/2023.iwslt-1.2",
    pages = "62--78",
}

@inproceedings{jiang-etal-2023-discourse,
      title="Discourse Centric Evaluation of Machine Translation with a Densely Annotated Parallel Corpus", 
      author="Yuchen Eleanor Jiang and Tianyu Liu and Shuming Ma and Dongdong Zhang and Ryan Cotterell and Mrinmaya Sachan",
      booktitle = "Proceedings of the 2023 Conference of the Association for Computational Linguistics: Human Language Technologies",
      month = jul,
      year = "2023",
      address = "Toronto, Canada",
      publisher = "Association for Computational Linguistics",
      url = "https://aclanthology.org/2023.acl-main.111",
      doi = "10.18653/v1/2023.main.111",
      pages = "1550--1565",
}

@inproceedings{wang2024findings,
  title={Findings of the WMT 2024 Shared Task on Discourse-Level Literary Translation},
  author={Wang, Longyue and Liu, Siyou and Wu, Minghao and Jiao, Wenxiang and Wang, Xing and Xu, Jiahao and Tu, Zhaopeng and Zhou, Liting and Gu, Yan and Chen, Weiyu and Koehn, Philipp and Way, Andy and Yuan, Yulin},
  booktitle={Proceedings of the Ninth Conference on Machine Translation},
  year={2024}
}

@article{wu2024adapting,
  title={Adapting large language models for document-level machine translation},
  author={Wu, Minghao and Vu, Thuy-Trang and Qu, Lizhen and Foster, George and Haffari, Gholamreza},
  journal={arXiv preprint arXiv:2401.06468},
  year={2024}
}

@inproceedings{ramos2025multilingual,
title={Multilingual Contextualization of Large Language Models for Document-Level Machine Translation},
author={Miguel Moura Ramos and Patrick Fernandes and Sweta Agrawal and Andre Martins},
booktitle={Second Conference on Language Modeling},
year={2025},
url={https://openreview.net/forum?id=Ah0U1r5Ldq}
}

@article{pang2025salute,
    author = {Pang, Jianhui and Ye, Fanghua and Wong, Derek Fai and Yu, Dian and Shi, Shuming and Tu, Zhaopeng and Wang, Longyue},
    title = {Salute the Classic: Revisiting Challenges of Machine Translation in
                    the Age of Large Language Models},
    journal = {Transactions of the Association for Computational Linguistics},
    volume = {13},
    pages = {73-95},
    year = {2025},
}

@inproceedings{wang-etal-2024-benchmarking,
    title = "Benchmarking and Improving Long-Text Translation with Large Language Models",
    author = "Wang, Longyue  and
      Du, Zefeng  and
      Jiao, Wenxiang  and
      Lyu, Chenyang  and
      Pang, Jianhui  and
      Cui, Leyang  and
      Song, Kaiqiang  and
      Wong, Derek  and
      Shi, Shuming  and
      Tu, Zhaopeng",
    editor = "Ku, Lun-Wei  and
      Martins, Andre  and
      Srikumar, Vivek",
    booktitle = "Findings of the Association for Computational Linguistics: ACL 2024",
    month = aug,
    year = "2024",
    address = "Bangkok, Thailand",
    publisher = "Association for Computational Linguistics",
    url = "https://aclanthology.org/2024.findings-acl.428/",
    doi = "10.18653/v1/2024.findings-acl.428",
    pages = "7175--7187",
}

@inproceedings{cui-etal-2024-efficiently,
    title = "Efficiently Exploring Large Language Models for Document-Level Machine Translation with In-context Learning",
    author = "Cui, Menglong  and
      Du, Jiangcun  and
      Zhu, Shaolin  and
      Xiong, Deyi",
    editor = "Ku, Lun-Wei  and
      Martins, Andre  and
      Srikumar, Vivek",
    booktitle = "Findings of the Association for Computational Linguistics: ACL 2024",
    month = aug,
    year = "2024",
    address = "Bangkok, Thailand",
    publisher = "Association for Computational Linguistics",
    url = "https://aclanthology.org/2024.findings-acl.646/",
    doi = "10.18653/v1/2024.findings-acl.646",
    pages = "10885--10897",
}

@inproceedings{lyu-etal-2024-paradigm,
    title = "A Paradigm Shift: The Future of Machine Translation Lies with Large Language Models",
    author = "Lyu, Chenyang  and
      Du, Zefeng  and
      Xu, Jitao  and
      Duan, Yitao  and
      Wu, Minghao  and
      Lynn, Teresa  and
      Aji, Alham Fikri  and
      Wong, Derek F.  and
      Wang, Longyue",
    editor = "Calzolari, Nicoletta  and
      Kan, Min-Yen  and
      Hoste, Veronique  and
      Lenci, Alessandro  and
      Sakti, Sakriani  and
      Xue, Nianwen",
    booktitle = "Proceedings of the 2024 Joint International Conference on Computational Linguistics, Language Resources and Evaluation (LREC-COLING 2024)",
    month = may,
    year = "2024",
    address = "Torino, Italia",
    publisher = "ELRA and ICCL",
    url = "https://aclanthology.org/2024.lrec-main.120/",
    pages = "1339--1352",
}

@inproceedings{xu2024a,
title={A Paradigm Shift in Machine Translation: Boosting Translation Performance of Large Language Models},
author={Haoran Xu and Young Jin Kim and Amr Sharaf and Hany Hassan Awadalla},
booktitle={The Twelfth International Conference on Learning Representations},
year={2024},
url={https://openreview.net/forum?id=farT6XXntP}
}

@inproceedings{li-etal-2025-enhancing-large,
    title = "Enhancing Large Language Models for Document-Level Translation Post-Editing Using Monolingual Data",
    author = "Li, Zongyao  and
      Rao, Zhiqiang  and
      Shang, Hengchao  and
      Guo, Jiaxin  and
      Li, Shaojun  and
      Wei, Daimeng  and
      Yang, Hao",
    editor = "Rambow, Owen  and
      Wanner, Leo  and
      Apidianaki, Marianna  and
      Al-Khalifa, Hend  and
      Eugenio, Barbara Di  and
      Schockaert, Steven",
    booktitle = "Proceedings of the 31st International Conference on Computational Linguistics",
    month = jan,
    year = "2025",
    address = "Abu Dhabi, UAE",
    publisher = "Association for Computational Linguistics",
    url = "https://aclanthology.org/2025.coling-main.591/",
    pages = "8830--8840",
}

@inproceedings{wang-etal-2023-document-level,
    title = "Document-Level Machine Translation with Large Language Models",
    author = "Wang, Longyue  and
      Lyu, Chenyang  and
      Ji, Tianbo  and
      Zhang, Zhirui  and
      Yu, Dian  and
      Shi, Shuming  and
      Tu, Zhaopeng",
    editor = "Bouamor, Houda  and
      Pino, Juan  and
      Bali, Kalika",
    booktitle = "Proceedings of the 2023 Conference on Empirical Methods in Natural Language Processing",
    month = dec,
    year = "2023",
    address = "Singapore",
    publisher = "Association for Computational Linguistics",
    url = "https://aclanthology.org/2023.emnlp-main.1036/",
    doi = "10.18653/v1/2023.emnlp-main.1036",
    pages = "16646--16661",
}
\clearpage
\appendix

\section{Prompt Templates}
\label{app:prompt_templates}

\begin{lstlisting}[caption=Document chunking prompt template., float=*, label=listing-chunk-prompt]
You are a document chunker. Here is an (incomplete) part of a long document:
{chunk_content}

Split the given text into contiguous, sentence-preserving chunks with coherent, adjacent sentences only. Remember to: (i) never split a sentence, (ii) keep topically related, adjacent sentences together, and (iii) if the last sentence is incomplete, do NOT output it - mark it as carry_over.

Return ONLY valid JSON with this schema:
{
  "chunks": [
    {
      "chunk_id": "<int|string>",
      "rationale": "<2 short sentences on why you choose the following sentences for this chunk>"
      "sentence_indices": [<ints, in order>]
      "carry_over": <true|false>,
    }
  ],...
}
\end{lstlisting}

\begin{lstlisting}[caption=Relation identification prompt template., float=*, label=listing-relation-identification]
You are a text relation analyzer. Your task is to examine two chunks of text, taken from the same document, **Chunk {i}** and **Chunk {j}** and determine if they share any meaningful connection. Only define a relation when there truly is one; if the two chunks are topically or semantically independent, indicate that explicitly.  

### Instructions

1. **Identify a Relation (if any)**  
   - Read **Chunk {j}** (the later chunk) and **Chunk {i}** (an earlier chunk).  
   - Ask yourself:  
     - Do they refer to the same entities, events, or ideas?  
     - Does Chunk {j} expand, clarify, contrast, or provide background for something in Chunk {i}?  
     - Does one chunk continue a thought begun in the other (e.g., cause -> effect, premise -> conclusion)?  
   - If there is an explicit connection, choose the most precise descriptor from the following categories:  
     - **Background->Core** ("Chunk {i} provides necessary background for Chunk {j}")
     - **Core->Detail** ("Chunk {j} drills down into a specific subpoint that Chunk {i} introduced")
     - **Motivation->Method or Problem->Solution** ("Chunk {i} states a problem; Chunk {j} describes the solution.")
     - **Cause->Effect** (e.g., "Chunk {i} describes an experiment; Chunk {j} shows the resulting performance drop.")
     - **Contrast** (e.g., "Chunk {j} explicitly contrasts with Chunk {i}'s claim.")
     - **Comparison** (e.g., "Both chunks compare two architectures.")
     - **Condition** (e.g., "Chunk {i} says 'if X, then Y,' and Chunk {j} describes what happens under that condition.")
     - **Evaluation** (e.g., "Chunk {j} evaluates the approach introduced in Chunk {i}.")
     - **Entity Coreference** ("Both chunks mention the same dataset, model, or variable.")
     - **Terminology Definition** ("Chunk {i} defines a term that appears in Chunk {j}.")

2. **Output Format**  
   - Output strictly a dictionary with keys: reason, relation, direction. Reason is a short explanation for the relation. If the relation is 'none', reason should be 'no relation found'. Direction is the direction of the relation, which can be 'forward' or 'backward'.
   - **Do not** include any extra commentary, annotations, or keys. Only the dictionary should be returned.

# Chunk {i}:
{chunk_i}
# Chunk {j}:
{chunk_j}

Your response:
\end{lstlisting}

\begin{lstlisting}[caption=Iterative chunk translation prompt template., float=*, label=listing-chunk-translation]
You are a high-quality translation assistant. Your task is to translate one specific chunk of the source document from {src_lang} into {tgt_lang}, using the following context to ensure consistent terminology, style, and meaning.

A. RELATED CHUNKS (and WHY THEY MATTER)

Below are all chunks that share a meaningful connection with the "Current Chunk."  
For each related chunk, you have:
  - Its "Chunk ID" (an integer).  
  - The full source-language text of that chunk.  
  - The detected relation-type between that chunk and the current one.  
  - A brief "Reason" sentence explaining why this relation helps guide terminology or meaning in the current chunk.  

Use these related-chunk definitions to preserve consistent translations of any key terms or ideas that overlap.

{related_chunks}

B. CURRENT CHUNK TO TRANSLATE

Chunk ID: {chunk_id}

Source Text:
{chunk_text}

C. SPECIFIC INSTRUCTIONS

1. **Consistent Terminology:**  
   - If any key term has been introduced or defined in a Related Chunk, use **exactly the same target-language rendering** here.  
   - If a Related Chunk indicates an acronym expansion, be sure to translate the acronym and its expanded form consistently (following how it was rendered earlier).

2. **Preserve Coreference & Referential Integrity:**  
   - If the Current Chunk refers back to an entity or concept defined earlier, ensure you use the same translation for that concept-exactly as in the Related Chunk's translation.

3. **Translate Only the Current Chunk:**  
   - Do not attempt to retranslate the entire document or other chunks.  
   - Your output should be **only** the translated text of {chunk_id}.  
   - Do **not** include any commentary, footnotes, or explanations-just the final translation block.

4. **Formatting:**  
   - Keep paragraph breaks as in the source.  
   - If the source chunk has multiple paragraphs, translate each paragraph and preserve line breaks.  
   - If the source chunk contains inline code, variable names, or labels, keep them in-English or code-style without adding extra formatting.


D. OUTPUT

Provide **only** the translated text of Chunk {chunk_id} in {tgt_lang}, respecting all the instructions above.
\end{lstlisting}

\begin{lstlisting}[caption=Prompt template for instructing LLM to annotate the reference words in the source text, float=*, label=listing-coreference-ann-prompt]
# Source Pronoun Annotation
{...} Role definition is omitted due to space limitations
## Task Definition
**Pronouns** are words that refer to entities mentioned elsewhere in the text or understood from context, including:
- **Personal pronouns**: I, you, he, she, it, we, they, me, him, her, us, them
- **Possessive pronouns**: my, your, his, her, its, our, their, mine, yours, hers, ours, theirs
- **Demonstrative pronouns**: this, that, these, those (when used as pronouns, not determiners)
- **Reflexive pronouns**: myself, yourself, himself, herself, itself, ourselves, yourselves, themselves
- **Relative pronouns**: who, whom, whose, which, that (when introducing relative clauses)

## Annotation Instructions

### Step 1: Read the entire document
Understand the content, identify all entities, and track referential relationships throughout the text.

### Step 2: Identify ALL pronouns
Scan systematically through the document for every pronoun that refers to a specific entity or concept.

### Step 3: Determine pronoun type and referent
- Classify the grammatical type of each pronoun
- Identify what specific entity or concept each pronoun refers to

### Step 4: Apply inline annotation
First copy each pronoun exactly as it appears in the original text, then add the attributes after it.

## Annotation Format
Use this exact format:
```
[pronoun]<type="[pronoun_type]" referent="[what_it_refers_to]">
```

**Attribute specifications:**
- `type`: One of: `personal`, `possessive`, `demonstrative`, `reflexive`, `relative`
- `referent`: The specific noun phrase or concept that this pronoun refers to

## Pronoun Type Guidelines

- **personal**: I, you, he, she, it, we, they, me, him, her, us, them
- **possessive**: my, your, his, her, its, our, their, mine, yours, hers, ours, theirs
{...} more types are omitted due to space limitations


## Examples

### Example 1 (Basic Pronouns):

**Source document**:
```
John bought a new car yesterday. He drove it to work this morning. Mary saw him and thought the car was beautiful. She told him that she liked it very much.
```

**Expected Output**:
```
John bought a new car yesterday. [He]<type="personal" referent="John"> drove [it]<type="personal" referent="a new car"> to work this morning. Mary saw [him]<type="personal" referent="John"> and thought the car was beautiful. [She]<type="personal" referent="Mary"> told [him]<type="personal" referent="John"> that [she]<type="personal" referent="Mary"> liked [it]<type="personal" referent="the car"> very much.
```
{...} more examples are omitted due to space limitations
\end{lstlisting}
\begin{lstlisting}[caption=Prompt template for conjunction annotation, float=*, label=listing-conjunction-ann-prompt]
# Source Conjunction Annotation
{...} Role definition is omitted due to space limitations
## Task Definition
**Conjunctive expressions** are words or phrases that signal logical relationships between clauses or sentences, including:
- **Coordinating conjunctions**: and, but, or, nor, for, so, yet
- **Subordinating conjunctions**: because, since, although, while, if, when, before, after, unless, etc.
- **Conjunctive adverbs**: however, therefore, furthermore, meanwhile, consequently, nevertheless, moreover, etc.
- **Transitional phrases**: in addition, on the other hand, as a result, for example, in contrast, etc.
- **Correlative conjunctions**: both...and, either...or, not only...but also, etc.

## Annotation Instructions
### Step 1: Read the entire document
Understand the content and identify the logical flow and relationships between clauses and sentences.

### Step 2: Identify ALL conjunctive expressions
Scan systematically through the document for every word or phrase that connects ideas or shows logical relationships.

### Step 3: Determine conjunction type and logical relationship
- Classify the grammatical type of each conjunctive expression
- Identify what type of logical connection it signals

### Step 4: Apply inline annotation
First copy each conjunctive expression exactly as it appears in the original text, then add the attributes after it.

## Annotation Format
Use this exact format:
```
[conjunction]<type="[conjunction_type]" relationship="[logical_relationship]">
```

**Attribute specifications:**
- `type`: One of: `coordinating`, `subordinating`, `conjunctive_adverb`, `transitional_phrase`, `correlative`
- `relationship`: One of the logical relationship categories listed below

## Logical Relationship Categories
- **addition**: adding information (and, furthermore, moreover, in addition)
{...} more categories are omitted due to space limitations

## Conjunction Type Guidelines

- **coordinating**: and, but, or, nor, for, so, yet
- **subordinating**: because, since, although, while, if, when, before, after, unless, though, whereas, etc.
{...} more types are omitted due to space limitations

## Examples

### Example 1 (Basic Conjunctions):

**Source document**:
```
John was tired, but he had to continue working. Therefore, he decided to have a cup of coffee. He drank it quickly and felt more energetic.
```
**Expected Output**:
```
John was tired, [but]<type="coordinating" relationship="contrast"> he had to continue working. [Therefore]<type="conjunctive_adverb" relationship="result">, he decided to have a cup of coffee. He drank it quickly [and]<type="coordinating" relationship="addition"> felt more energetic.
```
{...} more examples are omitted due to space limitations
\end{lstlisting}
\begin{lstlisting}[caption=Prompt template for instructing LLM to evaluate the reference cohensiveness, float=*, label=listing-coreference-eval-prompt]
# Reference Cohesion Translation Quality Evaluation
{...} Role definition is omitted due to space limitations
## Task Overview

You will receive an English source text that has been pre-annotated with pronoun information, along with a translation in a target language. Your job is to evaluate each annotated pronoun by determining:
1. How it was translated in the target language
2. Whether the translation is correct
3. If incorrect, what type of error occurred

## Evaluation Guidelines

For each annotated pronoun in the source text, you must add three new attributes to the existing annotation:

### Required Attributes to Add:
- **target_translation**: How the pronoun was rendered in the target language
- **is_correct**: Whether the translation is accurate (true/false)
- **error_type**: Type of error if translation is incorrect (null if correct)

### target_translation Values:
1. **Specific translation word(s)**: The actual translated pronoun (e.g., "Er", "elle")
2. **"omitted"**: The pronoun was appropriately omitted (common in pro-drop languages like Chinese/Japanese)
3. **"missing"**: The pronoun should have been translated but is absent

### is_correct Logic:
- If `target_translation` = specific word(s) -> `is_correct` can be true or false
- If `target_translation` = "omitted" -> `is_correct` must be true (appropriate omission)
- If `target_translation` = "missing" -> `is_correct` must be false (inappropriate absence)

### error_type Categories:
- **"null"**: No error (translation is correct)
- **"gender_mismatch"**: Wrong gender (he->she, him->her, etc.)
{...} more types are omitted due to space limitations

## Output Format

Return the complete annotated source text with the three new attributes added to each pronoun annotation:
```
[pronoun]<type="..." referent="..." target_translation="..." is_correct="true|false" error_type="...">
```

## Examples

### English Source Text
Tom and his sister went to the park. She found a ball and he picked it up. They decided to play together.
{...} English annotation is omitted due to space limitations
### German Translation  
**Translation**: Tom und seine Schwester gingen in den Park. Er fand einen Ball und er hob ihn auf. waren glucklich zusammen zu spielen.

**Expected Output**:
```
Tom and [his]<type="possessive" referent="Tom" target_translation="seine" is_correct="true" error_type="null"> sister went to the park. [She]<type="personal" referent="his sister" target_translation="Er" is_correct="false" error_type="wrong_referent"> found a ball and [he]<type="personal" referent="Tom" target_translation="er" is_correct="true" error_type="null"> picked [it]<type="personal" referent="a ball" target_translation="ihn" is_correct="true" error_type="null"> up. [They]<type="personal" referent="Tom and his sister" target_translation="missing" is_correct="false" error_type="missing_translation"> decided to play together.
```
{...} More examples are omitted due to space limitations
\end{lstlisting}
\begin{lstlisting}[caption=Prompt template for instructing LLM to evaluate the conjunction cohensiveness, float=*, label=listing-conjunction-eval-prompt]
# Conjunction Cohesion Translation Quality Evaluation
{...} Role definition is omitted due to space limitations
## Task Overview
You will receive an English source text that has been pre-annotated with conjunction information, along with a translation in a target language. Your job is to evaluate each annotated conjunction by determining:
1. How it was translated in the target language
2. Whether the translation preserves the correct logical relationship
3. If incorrect, what type of error occurred

## Evaluation Guidelines

For each annotated conjunction in the source text, you must add three new attributes to the existing annotation:

### Required Attributes to Add:
- **target_translation**: How the conjunction was rendered in the target language
- **is_correct**: Whether the translation preserves the logical relationship (true/false)
- **error_type**: Type of error if translation is incorrect (null if correct)

### target_translation Values:
1. **Specific translation word(s)**: The actual translated conjunction (e.g., "aber", "mais")
2. **"missing"**: The conjunction should have been translated but is absent

### is_correct Logic:
- If `target_translation` = specific word(s) -> `is_correct` can be true or false depending on logical relationship
- If `target_translation` = "missing" -> `is_correct` must be false (conjunction information lost)

### error_type Categories:
- **"null"**: No error (translation preserves correct logical relationship)
- **"wrong_conjunction"**: Conjunction translated but expresses wrong logical relationship
- **"missing_conjunction"**: Required conjunction is completely absent
- **"redundant_conjunction"**: Multiple conjunctions expressing same logical relationship
- **"inappropriate_addition"**: Adding conjunctions that create wrong logical relationships
- **"wrong_position"**: Conjunction in wrong syntactic position affecting meaning

## Output Format
Return the complete annotated source text with the three new attributes added to each conjunction annotation:
```
[conjunction]<type="..." relationship="..." target_translation="..." is_correct="true|false" error_type="...">
```

## Examples

### German Translation

**Annotated Source**:
{...} Source annotation is omitted due to space limitations

**Target Translation**: Das Wetter war schlecht, so entschieden wir uns zu wandern. Zuerst packten wir unsere Taschen. Dann verlieben wir fruh, weil wir den Verkehr vermeiden wollten. Obwohl es zu regnen begann, obwohl wir unsere Reise fortsetzten.

**Expected Output**:
```
The weather was bad, [but]<type="coordinating" relationship="contrast" target_translation="so" is_correct="false" error_type="wrong_conjunction"> we decided to go hiking. [First]<type="conjunctive_adverb" relationship="sequence" target_translation="Zuerst" is_correct="true" error_type="null">, we packed our bags. [Then]<type="conjunctive_adverb" relationship="sequence" target_translation="Dann" is_correct="true" error_type="null"> we left early [because]<type="subordinating" relationship="cause" target_translation="weil" is_correct="true" error_type="null"> we wanted to avoid traffic. [Although]<type="subordinating" relationship="concession" target_translation="Obwohl, obwohl" is_correct="false" error_type="redundant_conjunction"> it started raining, we continued our journey.
```
{...} More examples are omitted due to space limitations
\end{lstlisting}

We provide the prompt templates for \textsc{TransGraph} in this section. \Cref{listing-chunk-prompt} shows the prompt template for document chunking, \Cref{listing-relation-identification} shows the prompt template for relation labeling and finally, \Cref{listing-chunk-translation} is the prompt we used for iterative chunk translation.

Prompt templates used for evaluating the cohesion are also provided, including the annotation (\Cref{listing-coreference-ann-prompt}, \Cref{listing-conjunction-ann-prompt}) and evaluation (\Cref{listing-coreference-eval-prompt}, \Cref{listing-conjunction-eval-prompt}) on coreferences and conjunctions, respectively.

\section{Examples}

\begin{lstlisting}[caption=Example of document chunking., float=*, label=listing-chunk-example]
Chunk 1
Hello. My name is Asaf Harari. And I will present our paper, Few-Shot Tabular Data Enrichment Using Fine-Tuned Transformers Architectures. Data scientists analyze data and mainly focus on the manipulating the data's existing features. But sometimes, these features are limited. Feature generation using another data source may add substantial information. Our research goal is automatic tabular data enrichment using external sources' free text.

Chunk 2
Assume we have a tabular dataset and a knowledge base. We need an automatic process which involves entity linking and text analysis to extract new features from the knowledge base's free text. Our framework FeSTE is exactly this automatic process. So let's see an example in a dataset fed into FeSTE. In this example, the dataset is university dataset. When its goal is to classify universities into low ranking universities and high-ranking universities. As knowledge base, we use Wikipedia.

Chunk 3
The first phase of FeSTE is entity linking. When each entity, in this example the university name, is linked to an entity within the knowledge base. And and the text of the entities of the knowledge base is extracted and added to the dataset. In this example, the text is the Wikipedia page's abstract. Now, we need to generate or extract features from the retrieved text. So, we need to ah feature extraction phase ah which includes text analysis. And this is the main novelty of this paper and I will deep dive into it in the next slides.

Chunk 4
After the feature extraction phase, there is a feature generation phase when we use the extracted features to generate a small number of new features. First generate ah features in the number of classes of the original dataset. In this example, the original dataset has two classes. So, FeSTE generates two new features. But if the dataset has five classes, FeSTE generates five new features. Each feature represents the likelihood for each class. To analyze the text, we use the current state-of-the-art of text analysis, which are transformer based language models as BERT, GPT, XLNet and etc.

Chunk 5
It is but it is not likely that we can train language models using the input datasets. So a naive approach will be ah target task finetuning. So, in the feature extraction phase, we can download pretrained language models, finetune the language model over the target dataset. In this example to finetune the language model, to classify ah to classify text into classes, abstract into classes, low or high. Receive the language model output, which is the likelihood for each class and use as new features. The problem with this approach is datasets may have few distinct entities / texts. In our experiment, almost half of the datasets contain less than four hundred samples and the smallest dataset contain thirty five samples in its, in a training set. So to finetune a language model over ah this dataset will be ineffective.

...

Chunk 11
Here are the results for our experiments. You can see that we compare our our framework to target dataset finetuning, target task finetuning, and a MTDNN preliminary finetuning. And our reformulated finetuning achieves the best result, the best performance. While MTDNN achieved two percent improvement over the target dataset finetuning. Our approach achieved six percent improvement. When we look on the small ah dataset, we can see that the performance of MTDNN decreases and the improvement of the prelim, the preliminary multitask finetuning phase decreases to one point five percent. But our performance increased to eleven percent compared to the target task finetuning alone.

Chunk 12
For summing, FeSTE enables few shot enrichment from thirty five samples in our experiments. It uses one architecture for all tasks and datasets. And it keeps the head of ah of the model. But it adds reformulation phase. It augments the train set and it needs a target value with semantic meaning so we can feed it into the language model and use it in the sentence pair classification problem. Thank you.
\end{lstlisting}

\begin{table*}[t]
\renewcommand{\arraystretch}{1.5}
\centering
\footnotesize
\begin{tabular}{|l|p{5.5cm}|p{5.5cm}|}
\hline
\textbf{Relation} & \textbf{Chunk i} & \textbf{Chunk j} \\
\hline
Terminology Definition & Hello. My name is Asaf Harari. And I will present our paper, \textbf{\textcolor{blue}{Few-Shot Tabular Data Enrichment Using Fine-Tuned Transformers Architectures.}} & So let's see an example in a dataset fed into \textbf{\textcolor{blue}{FeSTE}}. In this example, the dataset is university dataset. When its goal is to classify universities into low ranking universities and high-ranking universities. As knowledge base, we use Wikipedia.
\\
\hline
Entity Coreference
&
So let's see an example in a dataset fed into \textbf{\textcolor{blue}{FeSTE}}. In this example, the dataset is university dataset. When its goal is to classify universities into low ranking universities and high-ranking universities. As knowledge base, we use Wikipedia.
&
The first phase of \textbf{\textcolor{blue}{FeSTE}} is entity linking. When each entity, in this example the university name, is linked to an entity within the knowledge base. And and the text of the entities of the knowledge base is extracted and added to the dataset. In this example, the text is the Wikipedia page's abstract.
\\
\hline
Background $\rightarrow$ Core
&
Data scientists analyze data and mainly focus on the manipulating the data's existing features. But sometimes, these features are limited. Feature generation using another data source may add substantial information. \textbf{\textcolor{blue}{Our research goal is automatic tabular data enrichment using external sources' free text.}}
&
Assume we have a tabular dataset and a knowledge base. We need an automatic process which involves entity linking and text analysis to extract new features from the knowledge base's free text. \textbf{\textcolor{blue}{Our framework FeSTE is exactly this automatic process.}}
\\
\hline
Core $\rightarrow$ Detail
&
Assume we have a tabular dataset and a knowledge base. We need an automatic process which involves entity linking and text analysis to extract new features from the knowledge base's free text. Our framework FeSTE is exactly this automatic process.
&
\textbf{\textcolor{blue}{So let's see an example in a dataset fed into FeSTE.}} In this example, the dataset is university dataset. When its goal is to classify universities into low ranking universities and high-ranking universities. As knowledge base, we use Wikipedia.
\\
\hline
Contrast
&
The state-of-the-art in multitask ah multitask finetuning called MTDNN. \textbf{\textcolor{blue}{In MTDNN, MTDNN maintains ah heads in the number of tasks in the training set}}. So, in this example there are four tasks in the training set, so MTDNN maintain four heads as you can see at the image. And it samples a random batch from ah from the training set. And if they random batch belongs to a, for example single sentence classification task, it executes forward and backward paths through the first head. And if the random batch belongs to pairwise ranking task, it executes forward and backward path through the last head.
&
In our scenario, ah tabular datasets vary in the number of classes. So there are many tasks. MTDNN maintained number of classes, heads, output layers. And the additional, additionally MTDNN needs to initialize new heads for a new dataset with a new task. \textbf{\textcolor{blue}{Our approach, called task reformulation finetuning is, in our approach task reformulation finetuning, instead of maintaining multiple heads, we reformulate each dataset into a sentence per classification problem, which is two classes' tasks.}}
\\
\hline
Comparison
&
\textbf{\textcolor{blue}{The state-of-the-art in multitask ah multitask finetuning called MTDNN. In MTDNN, MTDNN maintains ah heads in the number of tasks in the training set}}. So, in this example there are four tasks in the training set, so MTDNN maintain four heads as you can see at the image. And it samples a random batch from ah from the training set. And if they random batch belongs to a, for example single sentence classification task, it executes forward and backward paths through the first head. And if the random batch belongs to pairwise ranking task, it executes forward and backward path through the last head.
&
Here are the results for our experiments. \textbf{\textcolor{blue}{You can see that we compare our our framework to target dataset finetuning, target task finetuning, and a MTDNN preliminary finetuning. And our reformulated finetuning achieves the best result, the best performance. While MTDNN achieved two percent improvement over the target dataset finetuning}}. Our approach achieved six percent improvement. When we look on the small ah dataset, we can see that the performance of MTDNN decreases and the improvement of the prelim, the preliminary multitask finetuning phase decreases to one point five percent. But our performance increased to eleven percent compared to the target task finetuning alone.
\\
\hline
\end{tabular}
\caption{Relations between chunk pairs with blue-highlighted bold evidence provided by Qwen3-32B.}
\label{tab:relations-example}
\end{table*}

We provide the example of document chunking performed by Qwen3-32B in \Cref{listing-chunk-example}, while the examples for the relation identification process are given in \Cref{tab:relations-example}.

\end{document}